\documentclass[11pt]{article}

\usepackage[utf8]{inputenc}
\usepackage[T1]{fontenc}
\usepackage{times}
\usepackage[margin=1in]{geometry}
\usepackage{graphicx}
\usepackage{booktabs}
\usepackage{threeparttable}
\usepackage{array}
\usepackage{setspace}
\usepackage{caption}
\usepackage{amsmath}

\usepackage[hidelinks]{hyperref}
\usepackage{enumitem}  

\begin{document}

\title{Reducing Annotation Burden for Femoral Cartilage Segmentation in Knee MRI via Cross-Sequence Transfer Learning}
\date{}
\author{}
\maketitle

\begin{center}
{\large
Francesco Chiumento$^{1}$,
Gianluigi Crimi$^{2}$,
Elisa Moretta$^{2}$,
Rocco Milieri$^{3}$,
Alberto Bazzocchi$^{4}$,
Giulio Vara$^{5}$,
Giacomo Dal Fabbro$^{6}$,
Stefano Zaffagnini$^{6}$,
Fulvia Taddei$^{2,*}$,
Serena Bonaretti$^{7}$
}
\end{center}

\vspace{0.4cm}

\begin{flushleft}
\small
$^{1}$ School of Electronic Engineering, Dublin City University, Dublin, Ireland \\
$^{2}$ Bioengineering and Computing Laboratory, IRCCS Istituto Ortopedico Rizzoli, Bologna, Italy \\
$^{3}$ Dipartimento di Scienze Mediche e Chirurgiche (DIMEC), Alma Mater Studiorum -- Università di Bologna, Bologna, Italy \\
$^{4}$ Diagnostic and Interventional Radiology, IRCCS Istituto Ortopedico Rizzoli, Bologna, Italy \\
$^{5}$ Department of Biomedical and Neuromotor Sciences, University of Bologna, Bologna, Italy \\
$^{6}$ 2nd Orthopedics and Trauma Unit, IRCCS Istituto Ortopedico Rizzoli, Bologna, Italy \\
$^{7}$ Independent Researcher, Zurich, Switzerland \\
\vspace{0.2cm}
$^{*}$ Corresponding author: \href{mailto:fulvia.taddei@ior.it}{fulvia.taddei@ior.it}
\end{flushleft}

\vspace{0.4cm}

\newpage

\noindent\textbf{Summary Statement:}
Transfer learning enabled femoral cartilage segmentation across knee magnetic resonance imaging sequences with performance comparable to same-sequence training using fewer than 10 manually annotated target-domain subjects.

\section*{Key Points}
\begin{itemize}
    \item Transfer learning from sagittal proton density--weighted images (Cube) to dual-echo steady-state (DESS) images achieved segmentation performance comparable to same-sequence training (Dice similarity coefficient, 0.903 vs 0.900), reaching a performance plateau at 9 target-domain training subjects.
    \item Transfer was direction-dependent: Cube-to-DESS matched same-sequence performance with 9 target-domain subjects, whereas DESS-to-Cube remained below the Cube baseline (Dice similarity coefficient, 0.802 vs 0.830) and required 24 subjects to reach a plateau.
    \item Femoral cartilage lesions did not affect segmentation accuracy for DESS ($P \ge .39$) but significantly reduced accuracy for Cube (Dice similarity coefficient, 0.805 vs 0.856 in lesioned vs non-lesioned subjects; $P < .001$), suggesting that lesion-related performance degradation might be sequence-dependent.
\end{itemize}

\noindent\textbf{Abbreviations:}
ASD = average surface distance,
BMI = body mass index,
DESS = dual-echo steady-state,
DSC = Dice similarity coefficient,
KLG = Kellgren--Lawrence grade,
MRI = magnetic resonance imaging,
OAI = Osteoarthritis Initiative.

\newpage

\section*{Abstract}

\noindent\textbf{Purpose:} To develop and evaluate cross-sequence transfer learning for automatic femoral cartilage segmentation, testing bidirectional transfer between dual-echo steady-state (DESS) and sagittal proton density--weighted 3D fast spin-echo (Cube) sequences.

\vspace{0.3cm}

\noindent\textbf{Materials and Methods:}
We optimized a modified 2D U-Net on 507 DESS images from the Osteoarthritis Initiative (OAI). 
We then established same-sequence baselines using subject-level cross-validation on a subset of 44 OAI DESS images and 44 Cube images acquired at the Istituto Ortopedico Rizzoli, Bologna, Italy. Each subset included 22 non-lesioned and 22 lesioned subjects.
Finally, we performed transfer learning across sequences by fine-tuning the pretrained models on the target sequence with increasing training set sizes to study convergence, while keeping validation and test sets fixed. Segmentations were evaluated using Dice similarity coefficient (DSC) and average surface distance (ASD). Lesion effects were assessed with two-sided Mann--Whitney U tests with Bonferroni correction.

\vspace{0.3cm}

\noindent\textbf{Results:} Same-sequence training yielded higher accuracy on DESS than Cube (DSC, $0.900$ vs $0.830$; $P < .001$). Cube-to-DESS transfer matched DESS performance (DSC, $0.903 \pm 0.032$ vs $0.900 \pm 0.027$), reaching a performance plateau at 9 training subjects. DESS-to-Cube yielded a lower combined DSC ($0.802 \pm 0.049$ vs $0.830 \pm 0.042$), reaching a plateau at 24 training subjects. Lesions did not affect DESS ($P \ge .39$) but reduced Cube accuracy (DSC, $0.805$ vs $0.856$; $P < .001$).

\vspace{0.3cm}

\noindent\textbf{Conclusion:} Transfer learning across sequences can substantially reduce target-sequence annotation requirements for femoral cartilage segmentation, but performance is direction- and sequence-dependent, and the effects of lesions on segmentation may vary across MRI sequences.

\newpage

\section{Introduction}

The degeneration of femoral articular cartilage is a hallmark of knee osteoarthritis~\cite{hadaDegenerationDestructionFemoral2014}.
Cartilage changes are often quantified from magnetic resonance (MR) images, where cartilage is segmented to measure its composition and morphology~\cite{wangUseMagneticResonance2012,wirthQuantitativeMeasurementCartilage2023}.
Although historically segmentation has been a major obstacle in this workflow, in recent years, deep learning (DL) methods have demonstrated promising improvements in efficiency and accuracy~\cite{khanAdvancingDeepLearning2026}.
Most of the existing DL models in the literature are trained and evaluated on a single MR sequence \cite{panfilovDeepLearningbasedSegmentation2022a,normanUse2DUNet2018}, predominantly the dual-echo steady-state (DESS) from the Osteoarthritis Initiative (OAI)~\cite{peterfyOsteoarthritisInitiativeReport2008}.  
However, clinical practice may involve other acquisition sequences, such as proton density--weighted 3D fast spin-echo (e.g., Cube), which can provide different tissue contrast and spatial resolution \cite{friedrichHighresolutionCartilageImaging2011,tokudaMRIAnatomicalStructures2012}. 

Transfer learning is a DL technique where knowledge acquired from the source dataset is adapted (or fine-tuned) to a new target dataset~\cite{salehiStudyCNNTransfer2023}.
In MR imaging, directly transferring a model trained on one source sequence (e.g., DESS) to a new target sequence (e.g., Cube) often degrades performance.
To restore segmentation accuracy, the model typically needs to be retrained with a large number of labeled images from the target dataset~\cite{tajbakhshConvolutionalNeuralNetworks2016}.
However, these labels are often unavailable and their creation can require substantial effort by human annotators~\cite{yangAutomatedKneeCartilage2022}. 
Among transfer learning approaches, partial fine-tuning addresses this issue by updating only selective layers of a model with a small number of labeled target examples~\cite{ghafoorianTransferLearningDomain2017}.
To date, it remains unclear whether partial fine-tuning transfer learning can achieve performance comparable to same-sequence training for femoral cartilage segmentation and how many target-domain labeled images are needed to reach a stable performance plateau. 

In this study, we aim to:
\begin{enumerate}[noitemsep, topsep=0pt]
    \item Optimize a U-Net architecture for the segmentation of knee femoral cartilage to establish a control baseline, using DESS images.
    \item Separately evaluate the optimized model on a subset of DESS images and Cube images (same-sequence learning).
    \item Implement and evaluate transfer learning across MRI sequences by fine-tuning the model pretrained on DESS images for Cube images and vice versa (cross-sequence transfer learning).  
    In addition, we quantify the minimum number of target-domain labeled examples required to achieve performance comparable to same-sequence learning. 
\end{enumerate}
While evaluating same-sequence learning and cross-sequence transfer learning, we also assessed the effect of cartilage lesion status on segmentation accuracy by using three separate subject splits: non-lesioned, lesioned, and the combined cohort.

\section{Materials and Methods}

\subsection{Datasets}

In this retrospective study, we used knee MR images acquired with DESS and Cube protocols.
The DESS images were obtained from the OAI, a multicenter longitudinal observational study involving nearly 5,000 subjects (recruitment: February 2004--May 2006) \cite{peterfyOsteoarthritisInitiativeReport2008}. 
The Cube images were derived from a prospective study involving 49 subjects conducted at Istituto Ortopedico Rizzoli (Bologna, Italy) between March 2021 and May 2024~\cite{grenno2026high}.
Both studies received ethical approval and all participants provided informed consent.

Manual segmentations were available for both datasets: 507 for the DESS images (released by the Zuse Institute Berlin \cite{ambellanAutomatedSegmentationKnee2019}) and 44 for the Cube images, performed in-house by expert operators.
To optimize the U-Net architecture (aim~1), we used the full set of 507 images from the OAI dataset.
For the same-sequence learning (aim~2) and cross-sequence transfer learning (aim~3), we used DESS and Cube images from 44 segmented subjects (22 without femoral cartilage lesions and 22 with femoral cartilage lesions). Lesion status was defined by the presence of focal cartilage defects (partial- or 
full-thickness cartilage loss).
To obtain comparable cohorts, we matched subjects with DESS images to subjects with Cube images using a nearest-neighbor approach with Euclidean distance based on age, sex, body mass index (BMI), laterality, and Kellgren--Lawrence grade (Fig.~\ref{fig:workflow}, Phase~I, A).
Details of subject demographics and image characteristics are summarized in Table~\ref{tab:demographics}.

\begin{table}[!htbp]
\centering
\caption{Subject demographics and dataset characteristics.}
\label{tab:demographics}
\begin{threeparttable}
\begin{tabular}{lccccc}
\toprule
\textbf{Characteristic} 
& \multicolumn{1}{c}{\textbf{Optimization}} 
& \multicolumn{4}{c}{\textbf{Sequence learning}} \\
& \multicolumn{1}{c}{\textbf{DESS}} 
& \multicolumn{2}{c}{\textbf{DESS}} 
& \multicolumn{2}{c}{\textbf{Cube}} \\
\cmidrule(lr){2-2} \cmidrule(lr){3-4} \cmidrule(lr){5-6}
& 
& Non-Lesioned 
& Lesioned 
& Non-Lesioned 
& Lesioned \\
& ($n=507$) 
& ($n=22$) 
& ($n=22$) 
& ($n=22$) 
& ($n=22$) \\
\midrule
Age (years)         & 61.87 $\pm$ 9.33 & 62.6 $\pm$ 9.2  & 63.1 $\pm$ 9.2  & 50.9 $\pm$ 9.6  & 56.4 $\pm$ 7.1 \\
Sex (M/F)           & 262/245 & 12/10           & 19/3            & 17/5            & 18/4 \\
BMI (kg/m$^2$)      & 29.27 $\pm$ 4.52 & 27.9 $\pm$ 3.8  & 28.6 $\pm$ 4.3  & 26.1 $\pm$ 4.0  & 26.9 $\pm$ 3.6 \\
Laterality (L/R)$^{a}$ & 0/507 & 0/22         & 0/22            & 9/13            & 10/12 \\
KLG (0/1/2/3/4)$^{b}$   & 60/77/61/151/158 & 0/0/14/8/0          & 0/0/0/13/9          & 0/0/17/5/0          & 0/0/4/18/0 \\
Timepoint  & Baseline & Baseline & Baseline & Baseline & Baseline \\ 
\midrule
MRI Scanner         & \multicolumn{3}{c}{Siemens 3~T Trio} & \multicolumn{2}{c}{GE 3~T Discovery MR750} \\
MRI Sequence        & \multicolumn{3}{c}{DESS} & \multicolumn{2}{c}{Sagittal PD-weighted Cube} \\
Matrix              & \multicolumn{3}{c}{$384\times384\times160$} & \multicolumn{2}{c}{$512\times512\times296$--$312$} \\
Voxel Spacing (mm)  & \multicolumn{3}{c}{$0.3646\times0.3646\times0.7000$} & \multicolumn{2}{c}{$0.4121\times0.4121\times0.4000$} \\
\bottomrule
\end{tabular}
\begin{tablenotes}
\footnotesize
\item $^{a}$ Right knees were flipped to left orientation during preprocessing.
\item $^{b}$ In sequence learning, KLG $\geq$ 2 in all cohorts; no KLG 0 or KLG 4 subjects were available in the Cube cohort for matching.
\end{tablenotes}
\end{threeparttable}
\end{table}

\subsection{Image Preprocessing and 2D Dataset Construction}

Preprocessing included uniform orientation alignment, laterality correction (flipping right knees to left), registration to a common origin, bias field correction, intensity rescaling with cartilage contour enhancement, and anisotropic diffusion-based filtering~\cite{bonarettiPyKNEErImageAnalysis2020}. 
In addition, we converted each preprocessed 3D volume into 2D sagittal slices with corresponding binary masks for network training (Fig.~\ref{fig:workflow}, Phase~I, B).

\subsection{U-Net architecture}
\begin{figure}[!h]
    \centering
    \includegraphics[width=\textwidth]{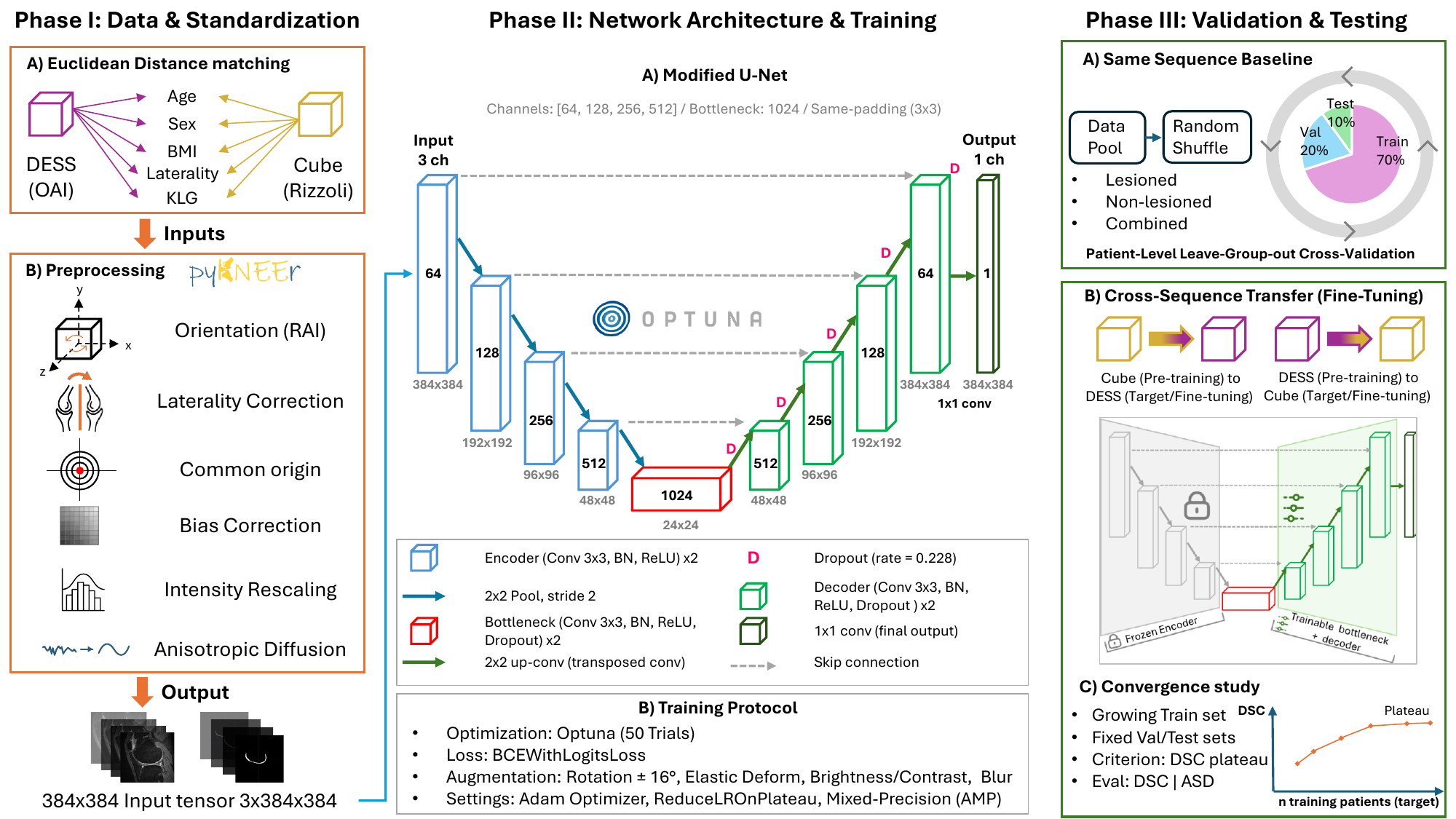}
    \caption{Study workflow including cohort matching (Phase~I, A), standardized preprocessing (Phase~I, B), modified U-Net architecture (Phase~II, A, B), subject-level cross-validation, and cross-sequence fine-tuning with convergence analysis (Phase~III, A, B, C).}
    \label{fig:workflow}
\end{figure}

We implemented a 2D U-Net architecture based on the encoder--decoder structure proposed by Ronneberger et al.~\cite{ronnebergerUNetConvolutionalNetworks2015} (Fig.~\ref{fig:workflow}, Phase~II, A). 
The encoder used four resolution levels (64, 128, 256, 512 channels), each with two $3\times3$ convolutions and $2\times2$ max pooling. 
The bottleneck used 1024 channels. 
The decoder used $2\times2$ transposed convolutions with skip connections, and a final $1\times1$ convolution produced a single-channel output. 
Compared with the original architecture, we replaced valid convolutions with same-padded convolutions to preserve input--output spatial dimensions. 
After each convolutional layer, we added batch normalization, and after each ReLU activation within the bottleneck and decoder blocks, we introduced dropout regularization.

\subsection{Optimizing model hyperparameters and data augmentation parameters}
\label{sec:exp1}

For the U-Net model, we optimized the loss function (choosing among binary cross-entropy with logits, Dice loss, generalized Dice loss, and focal loss), learning rate, batch size, and dropout rate.
We used a Tree-structured Parzen Estimator \cite{bergstraAlgorithmsHyperParameterOptimization2011} over 50 trials to maximize the validation Dice similarity coefficient (DSC).
To accelerate the search, we resized images to $192\times192$ pixels (Fig.~\ref{fig:workflow}, Phase~II, A, B).
For data augmentation, we optimized rotation, elastic deformation, brightness/contrast adjustment, and Gaussian blur. 
All the optimized values were used in the sequence learning experiments.
For cross-sequence transfer learning, we used a reduced learning rate and smaller batch size during fine-tuning.

\subsection{Same-sequence learning}
\label{sec:exp2}
We trained each model from scratch at native in-plane resolution for up to 100 epochs with the Adam optimizer and mixed-precision training, retaining the checkpoint with the highest validation DSC.
The learning rate was reduced on plateau starting from epoch 50 (factor $0.1$, patience $10$ epochs).  
We used subject-level cross-validation (70\% training, 20\% validation, 10\% test), 
ensuring that each subject appeared in the test set once across all iterations (Fig.~\ref{fig:workflow}, Phase~III, A).
During inference, we reconstructed 3D volumes by reversing the preprocessing steps (reorientation, padding, and cropping) and restoring the original metadata.

\subsection{Cross-sequence transfer learning}
\label{sec:exp3} 

During fine-tuning (up to 50 epochs), the learning rate was set to $1\times10^{-5}$ to preserve pretrained feature representations \cite{yosinskiHowTransferableAre2014,tajbakhshConvolutionalNeuralNetworks2016}, with a batch size of 4 (Fig.~\ref{fig:workflow}, Phase~III, B).
The encoder was frozen to preserve learned low-level features \cite{raghuTransfusionUnderstandingTransfer2019}, while the bottleneck and decoder were updated. 
The learning rate was reduced on plateau from epoch 10 (factor $0.5$, patience 5 epochs, minimum learning rate $10^{-7}$), and early stopping was used (patience 15 epochs, minimum improvement $\delta=0.001$). 
To determine the minimum number of target-domain subjects required to achieve performance comparable to same-sequence learning,
we conducted convergence studies (Fig.~\ref{fig:workflow}, Phase~III, C). 
We incrementally increased the training set from 1 to the maximum number of subjects available, while keeping validation set and test set fixed (non-lesioned and lesioned cohorts: 3 validation, 5 test; combined cohort: 5 validation, 8 test). 
Convergence was defined as an absolute improvement of less than 0.003 in DSC over four consecutive increments. 
During inference, we applied the same reconstruction procedure as in same-sequence learning.

\subsection{Segmentation evaluation and lesion-effect analysis}\label{sec:eval}

We evaluated segmentation performance using DSC $\left(\frac{2|P\cap G|}{|P|+|G|}\right)$, where $P$ and $G$ denote the predicted and reference masks, and ASD (mm)~\cite{tahaMetricsEvaluating3D2015}.
In addition, we compared segmentation performance between sequences (DESS vs Cube) in the combined cohort and assessed lesion-related performance differences within each sequence.
For lesion-status comparisons, we stratified per-subject test results from the combined-cohort experiment (lesioned vs non-lesioned). This approach avoids confounding from cross-model comparisons. Differences were assessed using two-sided Mann--Whitney U tests on DSC and ASD. 
We applied Bonferroni correction across two metrics ($\alpha_{\mathrm{corr}}=0.025$). 
We report unadjusted $P$ values and consider $P<\alpha_{\mathrm{corr}}$ significant.

\subsection{Computational details and reproducibility}
All code was written and executed in Python 3.12 using open-source packages.
We preprocessed all MRI data and computed segmentation metrics using pyKNEEr \cite{bonarettiPyKNEErImageAnalysis2020}.
To optimize hyperparameters and data augmentation parameters, we used Optuna 3.6.1 \cite{akibaOptunaNextgenerationHyperparameter2019}.
The U-Net models were implemented in PyTorch 2.7 with Albumentations 1.4 for data augmentation and trained on an NVIDIA GeForce RTX 4090. 
Statistical analyses were performed using SciPy 1.15.
Our code is available on GitHub at \url{https://github.com/FrancescoChiumento/femoral-cartilage-TL} where it is structured according to the guidelines of the Open and Reproducible Musculoskeletal Imaging Research (ORMIR) Community~\cite{bonaretti2026open}.

\section{Results}
\label{sec:results}
\subsection{Parameter optimization and data augmentation}

The optimal hyperparameter configuration was: binary cross-entropy with logits, learning rate $= 6.21\times10^{-5}$, batch size $= 7$, and dropout rate $= 0.228$. 
For data augmentation, the optimal parameters were: rotation $\pm16^\circ$ with probability $0.5$; elastic deformation with $\alpha=23.59$ and $\sigma=5.62$ with probability $0.5$; brightness/contrast adjustment with probability $0.128$; Gaussian blur with kernel size $7$--$11$ pixels with probability $0.1$.

\subsection{Quantitative results}

\paragraph{Same-sequence learning}
For DESS, same-sequence training yielded DSC $0.894 \pm 0.025$ (non-lesioned), $0.900 \pm 0.025$ (lesioned), and $0.900 \pm 0.027$ (combined). For Cube, DSC was $0.849 \pm 0.023$ (non-lesioned), $0.791 \pm 0.037$ (lesioned), and $0.830 \pm 0.042$ (combined). Full results are reported in Table~\ref{tab:results}. Performance was comparable across lesion status within DESS, whereas within Cube non-lesioned subjects achieved higher DSC than lesioned subjects.

\begin{figure}[!h]
\centering

\begin{minipage}{\textwidth}
    \centering

    \begin{minipage}[b]{0.3295\textwidth}
        \centering
        \includegraphics[width=\textwidth]{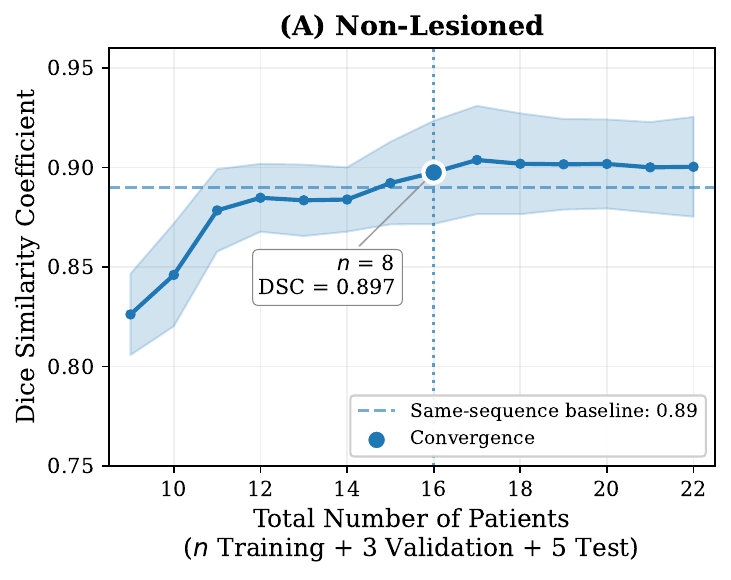}
    \end{minipage}
    \hfill
    \begin{minipage}[b]{0.3295\textwidth}
        \centering
        \includegraphics[width=\textwidth]{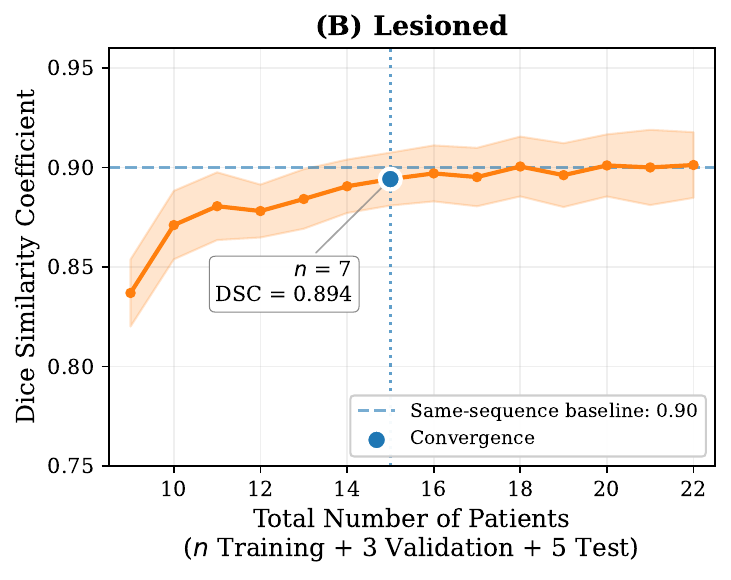}
    \end{minipage}
    \hfill
    \begin{minipage}[b]{0.3295\textwidth}
        \centering
        \includegraphics[width=\textwidth]{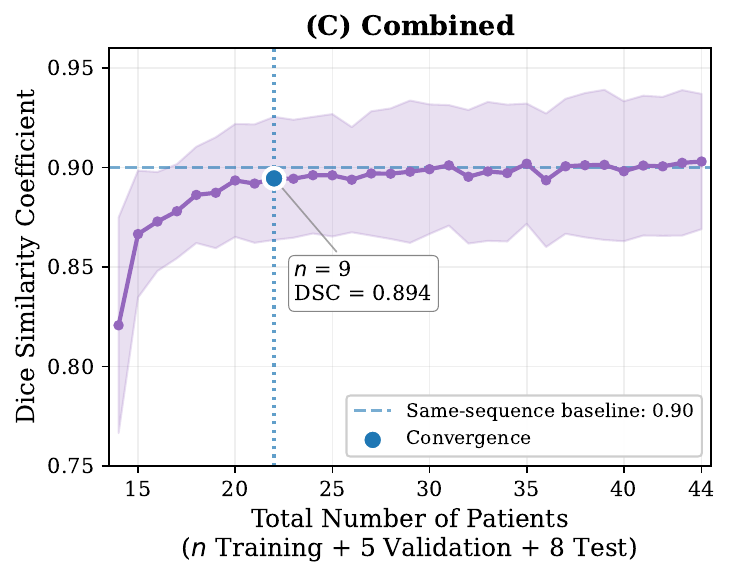}
    \end{minipage}

    \caption{Cube-to-DESS transfer convergence for (A) non-lesioned, (B) lesioned, (C) combined knees. 
    Markers: Dice similarity coefficient; 
    shaded area: $\pm$1 standard deviation; 
    dashed horizontal line: same-sequence performance; 
    dotted vertical line: convergence point (plateau training size). 
    }
    \label{fig:conv_cube_to_dess}
\end{minipage}

\vspace{0.6cm}

\begin{minipage}{\textwidth}
    \centering

    \begin{minipage}[b]{0.3295\textwidth}
        \centering
        \includegraphics[width=\textwidth]{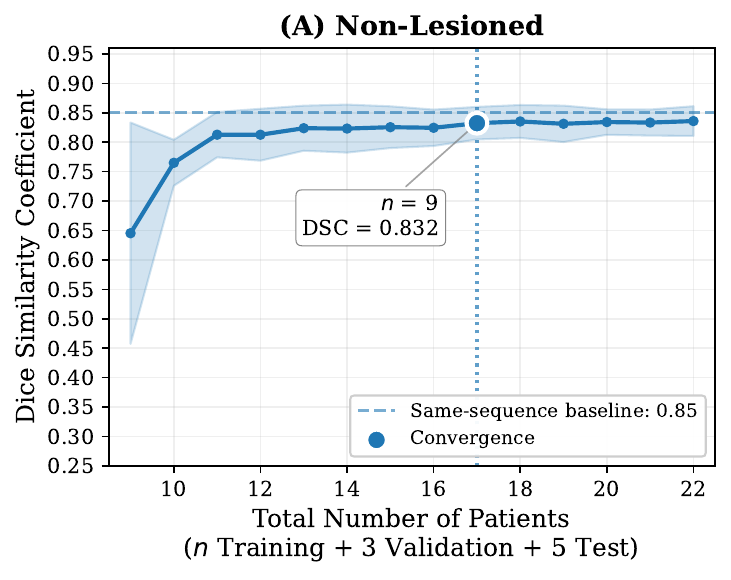}
    \end{minipage}
    \hfill
    \begin{minipage}[b]{0.3295\textwidth}
        \centering
        \includegraphics[width=\textwidth]{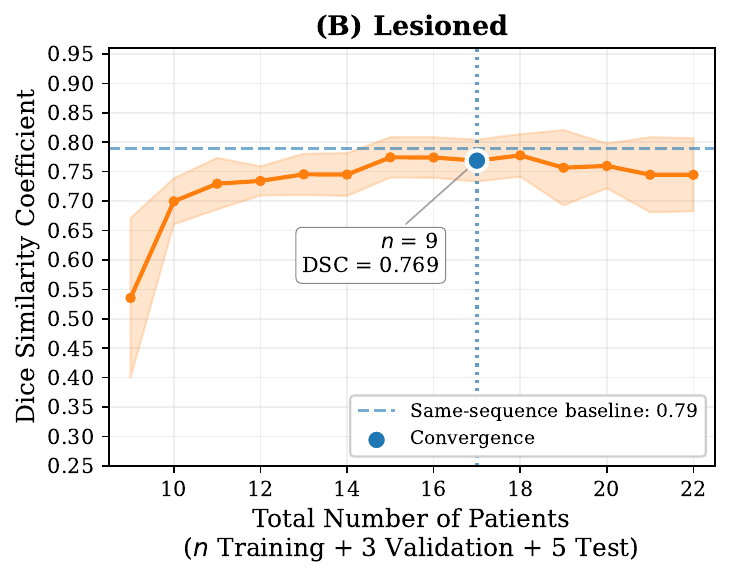}
    \end{minipage}
    \hfill
    \begin{minipage}[b]{0.3295\textwidth}
        \centering
        \includegraphics[width=\textwidth]{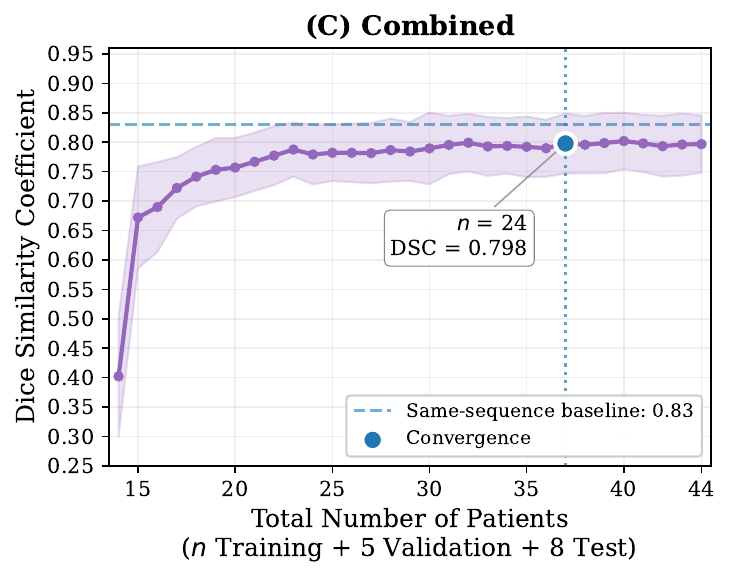}
    \end{minipage}

    \caption{DESS-to-Cube transfer convergence for (A) non-lesioned, (B) lesioned, and (C) combined. 
    Markers: Dice similarity coefficient; 
    shaded area: $\pm$1 standard deviation; 
    dashed horizontal line: same-sequence performance; 
    dotted vertical line: convergence point (plateau training size).
    }
    \label{fig:conv_dess_to_cube}
\end{minipage}

\end{figure}

\paragraph{Cross-sequence transfer learning}
Cube-to-DESS transfer learning achieved DSC $0.904 \pm 0.026$ (non-lesioned), $0.902 \pm 0.019$ (lesioned), and $0.903 \pm 0.032$ (combined), matching or exceeding same-sequence DESS performance. 
DESS-to-Cube transfer learning achieved DSC $0.834 \pm 0.024$ (non-lesioned), $0.776 \pm 0.036$ (lesioned), and $0.802 \pm 0.049$ (combined), below the same-sequence Cube baseline (Table~\ref{tab:results}). Cube-to-DESS reached a performance plateau at 8 (non-lesioned), 7 (lesioned), and 9 (combined) training subjects (Fig.~\ref{fig:conv_cube_to_dess}). 
DESS-to-Cube reached a plateau at 9 (non-lesioned), 9 (lesioned), and 24 (combined) training subjects (Fig.~\ref{fig:conv_dess_to_cube}).

\begin{table}[!h]
\centering
\caption{Results of femoral cartilage segmentation. Bold indicates best Dice similarity coefficient (DSC) and average surface distance (ASD) across experiments.}
\label{tab:results}
\setlength{\tabcolsep}{5.05pt}
\begin{tabular}{l @{\hskip 1pt} c @{\hskip 1pt} cc}
\toprule
\textbf{Experiment} & \textbf{n (test)$^{a}$} & \textbf{DSC} & \textbf{ASD (mm)} \\
\midrule
\multicolumn{4}{l}{\textit{Same-Sequence Training}} \\
\midrule
Non-Lesioned (DESS)     & 22 & 0.894 $\pm$ 0.025 & 0.463 $\pm$ 0.094 \\
Lesioned (DESS)         & 22 & 0.900 $\pm$ 0.025 & 0.506 $\pm$ 0.192 \\
Combined (DESS)         & 44 & 0.900 $\pm$ 0.027 & 0.453 $\pm$ 0.129 \\
Non-Lesioned (Cube)     & 22 & 0.849 $\pm$ 0.023 & 0.796 $\pm$ 0.139 \\
Lesioned (Cube)         & 22 & 0.791 $\pm$ 0.037 & 1.681 $\pm$ 1.925 \\
Combined (Cube)         & 44 & 0.830 $\pm$ 0.042 & 0.978 $\pm$ 0.351 \\
\midrule
\multicolumn{4}{l}{\textit{Cross-Sequence Transfer Learning}} \\
\midrule
Non-Lesioned (Cube-to-DESS)  & 5 & \textbf{0.904} $\pm$ 0.026 & 0.444 $\pm$ 0.108 \\
Lesioned (Cube-to-DESS)  & 5 & 0.902 $\pm$ 0.019 & 0.472 $\pm$ 0.143 \\
Combined (Cube-to-DESS)  & 8 & 0.903 $\pm$ 0.032 & \textbf{0.425} $\pm$ 0.152 \\
Non-Lesioned (DESS-to-Cube)  & 5 & 0.834 $\pm$ 0.024 & 0.846 $\pm$ 0.122 \\
Lesioned (DESS-to-Cube)  & 5 & 0.776 $\pm$ 0.036 & 1.344 $\pm$ 0.213 \\
Combined (DESS-to-Cube)  & 8 & 0.802 $\pm$ 0.049 & 1.312 $\pm$ 0.579 \\
\bottomrule
\end{tabular}

\vspace{0.1cm}
\footnotesize{$^{a}$Same-sequence: n (test) = subjects evaluated via subject-level cross-validation. Transfer learning: n (test) = test subjects. Convergence training sizes are reported in Figs.~\ref{fig:conv_cube_to_dess}--\ref{fig:conv_dess_to_cube}. Non-lesioned, lesioned, and combined experiments used separate splits.}

\end{table}

\subsection{Qualitative examples}

\begin{figure}[!htbp]
    \centering
    \includegraphics[width=\textwidth]{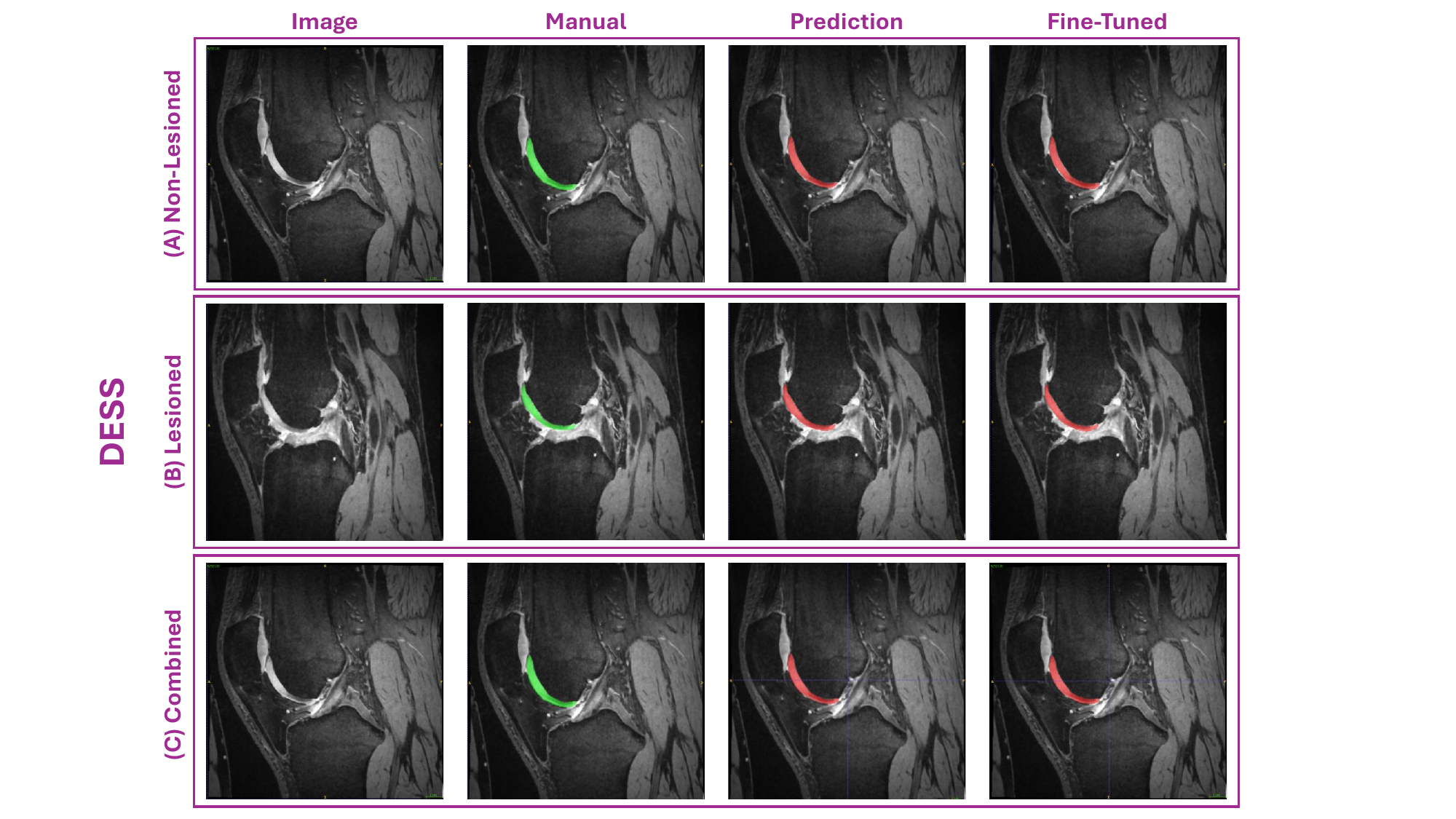}
    \captionsetup{skip=2pt}
    \caption{Qualitative examples of DESS segmentations. 
    Left to right: input; reference (green); same-sequence prediction (red); cross-sequence fine-tuned prediction (red). 
    Top to bottom: (A) non-lesioned; (B) lesioned; (C) combined.}
    \label{fig:dess_segmented}
\end{figure}

\begin{figure}[!htbp]
    \centering
    \includegraphics[width=\textwidth]{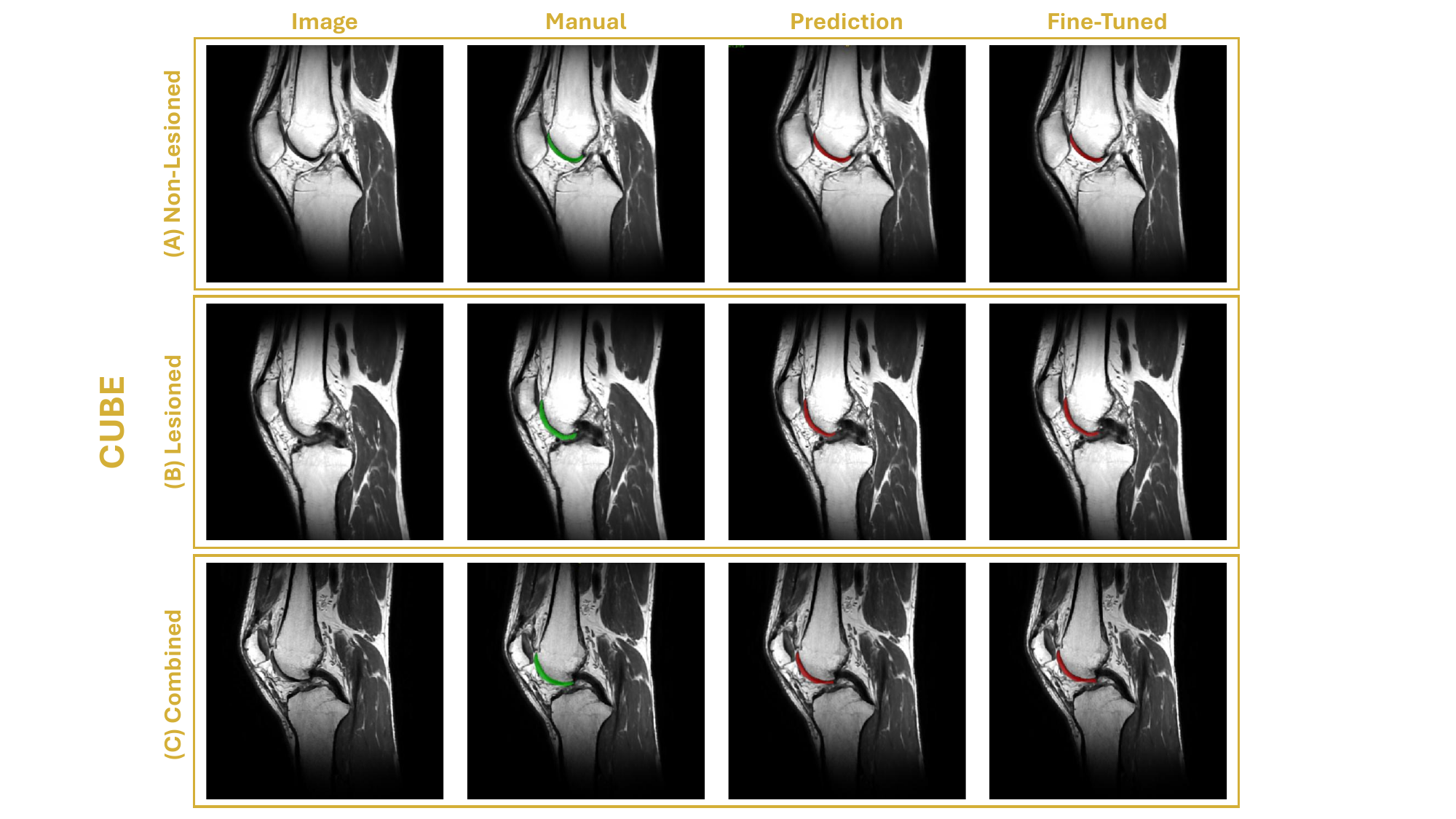}
    \captionsetup{skip=2pt}
    \caption{Qualitative examples of Cube segmentations. 
    Left to right: input; reference (green); same-sequence prediction (red); cross-sequence fine-tuned prediction (red). 
    Top to bottom: (A) non-lesioned; (B) lesioned; (C) combined.}
    \label{fig:cube_segmented}
\end{figure}

Representative segmentation examples are shown in Figures~\ref{fig:dess_segmented} and \ref{fig:cube_segmented}. Cube-to-DESS fine-tuning produced contours qualitatively comparable to same-sequence DESS across cases (Fig.~\ref{fig:dess_segmented}, Table~\ref{tab:results}). DESS-to-Cube fine-tuning improved overlap and boundary placement (Fig.~\ref{fig:cube_segmented}); however, some inaccuracies remained, more evident in lesioned subjects, consistent with the quantitative and statistical results.

\subsection{Statistical performance}

\begin{figure}[!htbp]
    \centering
    \includegraphics[width=\textwidth]{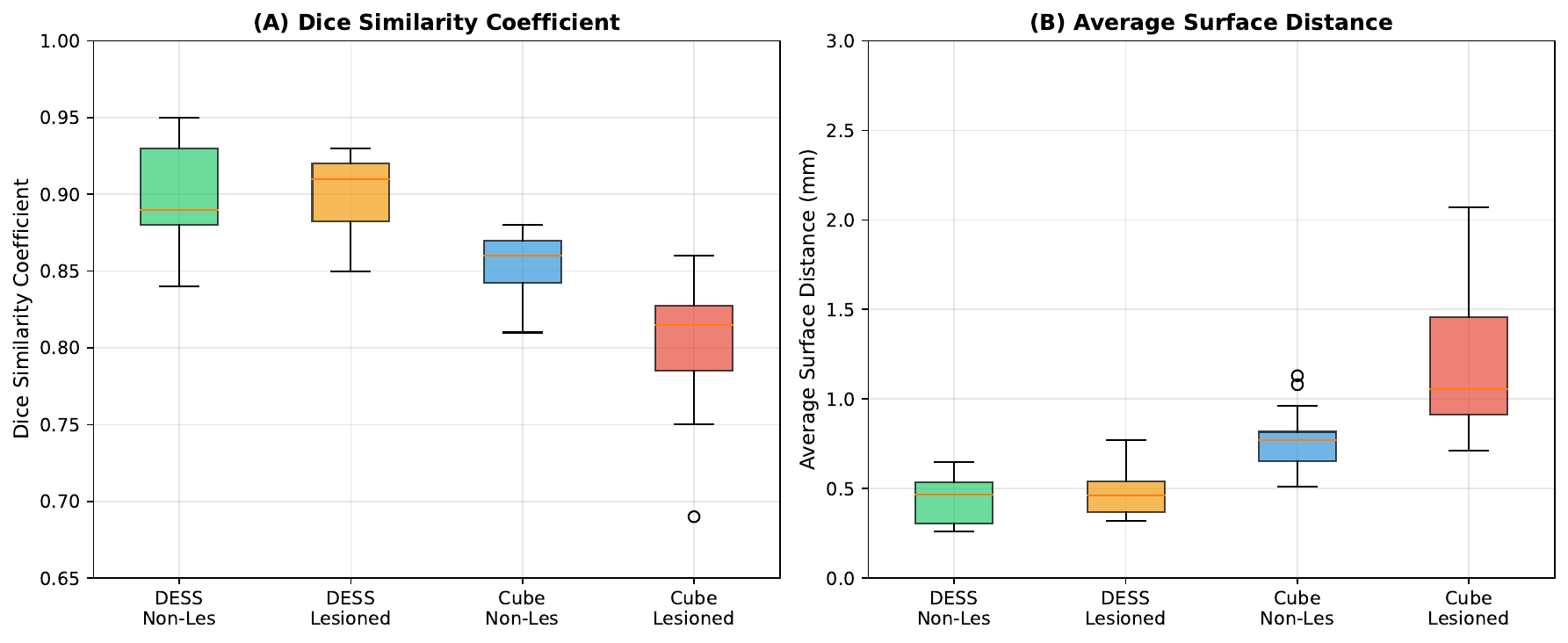}
    \caption{Subject-level box plots of (A) Dice similarity coefficient (DSC) and (B) average surface distance (ASD), stratified by sequence (DESS vs Cube) and lesion status (non-lesioned vs lesioned) within the combined-cohort experiment. Boxes show interquartile range/median; whiskers extend to $1.5{\times}$ the interquartile range; circles indicate outliers; ASD axis capped at 3.0~mm. Two-sided Mann--Whitney U with Bonferroni correction across two metrics ($\alpha_{\mathrm{corr}}=0.025$); $P$ unadjusted. DESS outperformed Cube in both DSC and ASD ($P < .001$). DESS: no lesion differences ($P \ge .39$ across metrics); Cube: lesioned worse in both DSC and ASD ($P < .001$).}
    \label{fig:boxplots}
\end{figure}

Across sequences (Fig.~\ref{fig:boxplots}), DESS showed higher DSC and lower ASD than Cube. 
Within Cube, lesioned subjects (DSC, $0.805 \pm 0.042$; ASD, $1.187 \pm 0.370$ mm) exhibited worse overlap and boundary accuracy than non-lesioned subjects (DSC, $0.856 \pm 0.021$; ASD, $0.770 \pm 0.157$ mm; both $P < .001$), whereas within DESS no significant lesion-related differences were observed ($P \ge .39$). These lesion-stratified values were computed within the combined-cohort experiment (Section~\ref{sec:eval}) and differ slightly from the lesion-specific models in Table~\ref{tab:results}.

\section{Discussion}

Same-sequence performance for DESS images (DSC, $0.900$) was comparable to prior OAI DESS benchmarks achieved with considerably larger training sets and more complex architectures. 
Norman et al.~\cite{normanUse2DUNet2018} reported a femoral cartilage DSC of $0.878$ using 174 DESS volumes with a 2D U-Net, and Gatti and Maly~\cite{gattiAutomaticKneeCartilage2021} reported $0.907$ on the OAI dataset using a computationally intensive multi-stage framework operating on 3D sub-volumes with 176 images from 88 subjects. 
Conversely, the performance for Cube was lower (DSC, $0.830$), likely reflecting the lower cartilage-to-background contrast of Cube~\cite{friedrichHighresolutionCartilageImaging2011}.

Transfer learning across sequences preserved segmentation quality while reducing annotation requirements.
The Cube-to-DESS transfer achieved comparable accuracy to same-sequence segmentation (DSC, $0.903$ vs $0.900$), whereas DESS-to-Cube transfer slightly underperformed the Cube baseline (DSC, $0.802$ vs $0.830$).
In addition, transfer learning performance was direction-dependent: Cube-to-DESS matched same-sequence performance with 9 target-domain subjects, whereas DESS-to-Cube required 24 subjects to reach a plateau. This asymmetry likely reflects the higher cartilage-to-tissue contrast of DESS~\cite{friedrichHighresolutionCartilageImaging2011}. Because the decoder and bottleneck learn to refine features for mask generation, fine-tuning these layers with well-contrasted data such as DESS may be sufficient for high-quality adaptation. In contrast, the lower inherent contrast of Cube may require more training data for effective knowledge transfer.

Lesion effects were sequence-dependent: DESS showed no lesion-related differences ($P \ge .39$); however, Cube performance decreased in lesioned subjects in the combined-cohort analysis (DSC, $0.805$ vs $0.856$; $P < .001$). DESS may better delineate damaged cartilage due to higher fluid-cartilage contrast \cite{friedrichHighresolutionCartilageImaging2011}, while Cube may be more susceptible when contrast is reduced~\cite{shahImagingUpdateCartilage2021}.

This study has some limitations. 
First, for computational reasons we evaluated a 2D U-Net with slice-wise inference followed by 3D reconstruction; future work should assess 3D architectures and additional domain-adaptation strategies. 
Second, these analyses focused on femoral cartilage; a natural extension will be to other knee structures, with the goal of expanding to multi-site targets to improve generalizability.
Finally, both datasets were acquired at $3$~T, and the generalizability of the method to $1.5$~T remains to be established.

In conclusion, cross-sequence transfer learning enabled femoral cartilage segmentation across MRI sequences, with Cube-to-DESS transfer achieving performance comparable to same-sequence training using fewer than $10$ annotated target-domain subjects. 
These findings suggest that transfer learning can reduce the manual annotation effort required to develop segmentation models across institution-specific MRI protocols.

\section*{Funding}
This study was partially funded by the Italian Ministry of Health (Project code: RF-2018-12368274).

\section*{Competing interests}
The authors have no competing interests to declare.

\section*{Acknowledgments}
We acknowledge the use of a large language model for editorial support, limited to grammar checking and language polishing. All scientific content, analyses, interpretation, and final text were reviewed and approved by the authors.

\section*{Author contributions statement}
Authors were asked to denote their contributions to the project according to the Contributor Roles Taxonomy (CRediT). Role abbreviations. C: Conceptualization, DC: Data curation for initial use and later re-use, FA: Funding acquisition, I: Investigation, M: Methodology, PA: Project administration, R: Resources, S: Software, SU: Supervision, V: Validation, W: Writing--original draft, and RE: Writing--review \& editing. \\

\noindent
Francesco Chiumento: C, DC, I, M, S, V, W, RE \\
Gianluigi Crimi: S, V \\
Elisa Moretta: DC, I, V \\
Rocco Milieri: I \\
Alberto Bazzocchi: M, SU \\
Giulio Vara: I \\
Giacomo Dal Fabbro: C, PA \\
Stefano Zaffagnini: FA, SU \\
Fulvia Taddei: C, FA, R, RE, SU \\
Serena Bonaretti: C, I, M, PA, R, SU, W, RE

\clearpage
\bibliographystyle{unsrt}
\bibliography{references.bib}

\end{document}